\documentclass{article}

\usepackage{microtype}
\usepackage{graphicx}
\usepackage{subfigure}
\usepackage{booktabs} 

\usepackage{hyperref}
\usepackage{makecell}
\usepackage{float}
\usepackage{xspace}

\newcommand{\name}{Vabs-Net\xspace}

\usepackage[accepted]{icml2024}

\usepackage{amsmath}
\usepackage{amssymb}
\usepackage{mathtools}
\usepackage{amsthm}
\usepackage{multirow}
\usepackage[capitalize,noabbrev]{cleveref}

\theoremstyle{plain}

\theoremstyle{definition}

\theoremstyle{remark}

\usepackage[textsize=tiny]{todonotes}

\icmltitlerunning{Pre-Training Protein Bi-level Representation Through Span Mask Strategy On 3D Protein Chains}

\begin{document}

\twocolumn[
\icmltitle{Pre-Training Protein Bi-level Representation Through Span Mask Strategy On 3D Protein Chains}



\icmlsetsymbol{equal}{*}

\begin{icmlauthorlist}
\icmlauthor{Jiale Zhao}{comp,sch}
\icmlauthor{Wanru Zhuang}{comp,sch3}
\icmlauthor{Jia Song}{comp,sch2}
\icmlauthor{Yaqi Li}{comp}
\icmlauthor{Shuqi Lu}{comp}
\end{icmlauthorlist}

\icmlaffiliation{sch}{Institute of Computing Technology,UCAS, Beijing, China}
\icmlaffiliation{comp}{DP Technology, Beijing, China}
\icmlaffiliation{sch2}{Xiamen University, Institute of Artificial Intelligence, Xiamen, China}
\icmlaffiliation{sch3}{Xiamen University, School of Informatics, Xiamen, China}

\icmlcorrespondingauthor{Shuqi Lu}{lusq@dp.tech}

\icmlkeywords{Machine Learning, ICML}

\vskip 0.3in
]



\printAffiliationsAndNotice{\icmlEqualContribution} 

\begin{abstract}
In recent years, there has been a surge in the development of 3D structure-based pre-trained protein models, representing a significant advancement over pre-trained protein language models in various downstream tasks. However, most existing structure-based pre-trained models primarily focus on the residue level, i.e., alpha carbon atoms, while ignoring other atoms like side chain atoms.
We argue that modeling proteins at both residue and atom levels is important since the side chain atoms can also be crucial for numerous downstream tasks, for example, molecular docking. Nevertheless, we find that naively combining residue and atom information during pre-training typically fails.
We identify a key reason is the information leakage caused by the inclusion of atom structure in the input, which renders residue-level pre-training tasks trivial and results in insufficiently expressive residue representations.
To address this issue, we introduce a span mask pre-training strategy on 3D protein chains to learn meaningful representations of both residues and atoms. This leads to a simple yet effective approach to learning protein representation suitable for diverse downstream tasks.  
Extensive experimental results on binding site prediction and function prediction tasks demonstrate our proposed pre-training approach significantly outperforms other methods. Our code will be made public.
\end{abstract}

\section{Introduction}
Protein modeling is essential, as proteins play critical roles in various cellular processes, such as transcription, translation, signaling, and cell cycle regulation. In recent years, advances in deep learning have significantly contributed to the development of pre-trained models for generating high-quality protein representations, enabling the prediction of diverse properties, like protein classification and function. Although many previous studies have focused on pre-training based on protein sequences~\cite{rao_msa_2021, lin2023evolutionary, chen_xtrimopglm_2023}, the significance of protein structure as a determinant of protein function has led to the emergence of 3D structure-based models~\cite{zhang2022protein, zhang2023pre, noauthor_pre-training_2023}. The development of these models has been greatly enhanced by recent breakthroughs in highly accurate deep learning-based protein structure prediction methods~\cite{jumper_highly_2021}. 

A common pre-training approach in previous studies leverages advances in self-prediction techniques within the field of natural language processing~\cite{brown2020language, devlin2018bert}. In this context, the objective for a given protein can be framed as predicting a specific segment of the protein based on the information from the remaining structure. Standard pre-training tasks often involve randomly masking certain residues, predicting the types and positions of the masked residues (i.e., the coordinates of alpha carbon atoms), and the angles between them and other residues~\cite{zhang2022protein, guo_self-supervised_2022, chen_structure-aware_2023}. Through this process, the model effectively captures the information of the residues, thereby obtaining a high-quality representation of the protein's residues.

\begin{figure}[ht]
    \centering
    \includegraphics[width=20pc]{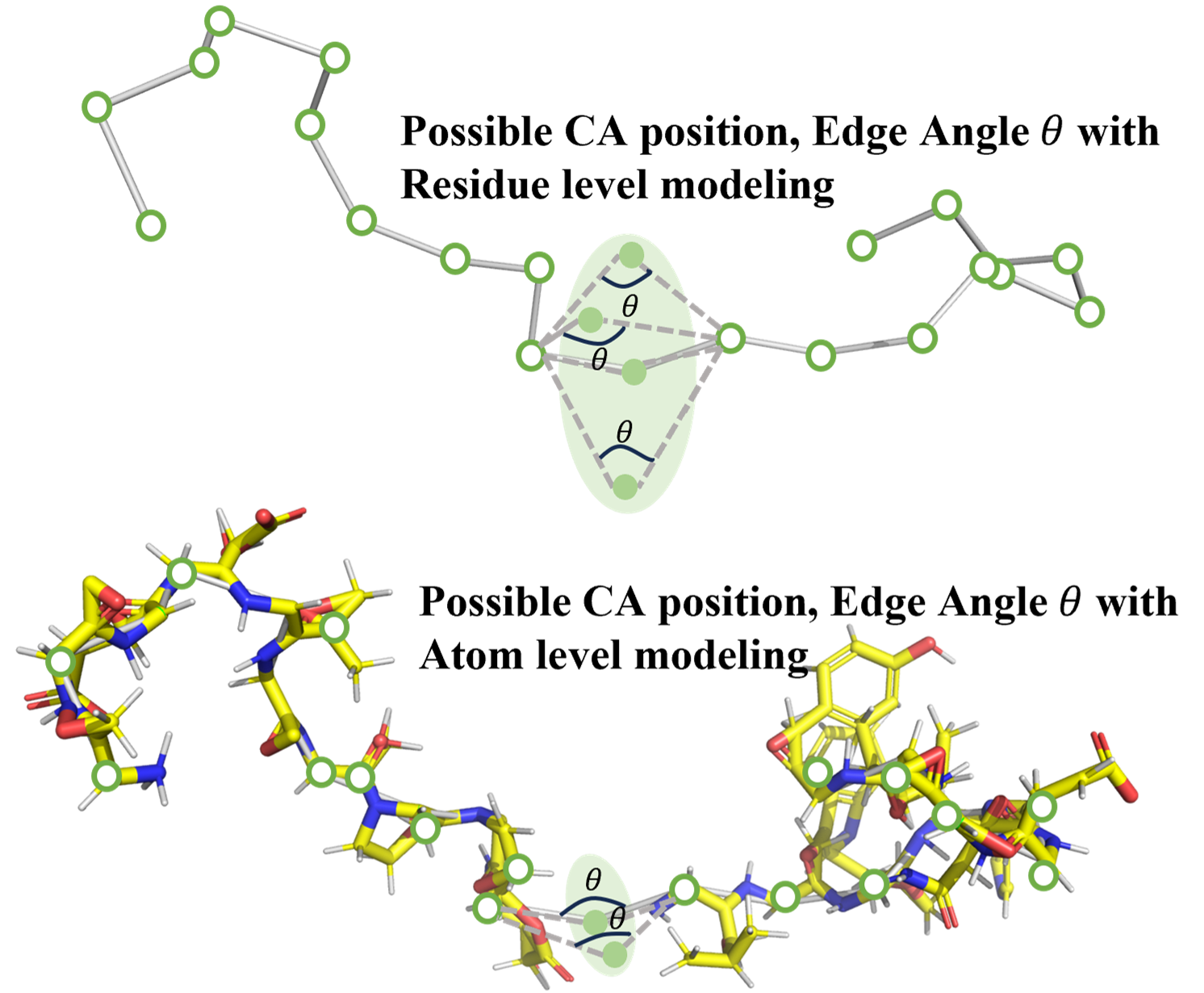}
    \caption{With all-atom added, the possible range for residue position is limited, thus resulting in easier prediction for residue position and angle between edges, etc. When all atoms are utilized, the prediction of residue positions and inter-edge angles relies predominantly on other atoms rather than residues themselves.
    }
    \label{fig:residue_atom_compare}
\end{figure}

Although many pre-trained models have demonstrated success in modeling the 3D structure of proteins, the majority focus on the residue level, utilizing only the geometric positions of alpha carbon atoms or main chain atoms. However, side chain atoms are also essential in numerous downstream tasks, such as molecular docking, due to their interactions with small molecules~\cite{krishna_generalized_2023}. Thus, it is imperative to incorporate information from all atoms in protein modeling.
In our empirical study, we find that naive atom-level modeling typically fails: (1) Simply replacing residue input with atom input, conducting pre-training tasks at the atom level without considering residue level, such as predicting atom coordinates and angles, does not yield significant improvements. This suggests that residue-based modeling is indispensable. (2) Naively combining residue and atom information and conducting pre-training tasks at both atom and residue levels does not lead to enhanced performance.
We identified a key reason to be the information leakage of residue-level tasks caused by the inclusion of atom structure in the input, which renders residue-level pre-training tasks trivial and results in insufficiently expressive residue representations. As shown in Figure~\ref{fig:residue_atom_compare}, the current residue structure can be easily predicted when surrounded by atom structure. This indicates the necessity of properly modeling residue information. To address this, we introduce a Span Mask strategy on 3D Protein Chains (SMPC). We mask the residue type of consecutive biologically meaningful substructures, retaining only the alpha carbon atom for the span-masked residues while eliminating all other atoms. This approach increases the difficulty of residue tasks, making it impossible to infer residue type and structure solely from side-chain and backbone atoms, thus, prompting the model to learn meaningful residue representations.

Our analysis leads to a simple yet effective approach. In this paper, we propose a \textbf{V}ector \textbf{A}ware \textbf{B}ilevel \textbf{S}parse Attention Network (\name), a pre-trained model that simultaneously models residues and atoms. \name employs a carefully designed edge vector encoding module and a two-track sparse attention module which comprise an atom-atom track and a residue-residue track, to encode atoms and residues. These tracks interact via alpha carbon atoms. To ensure meaningful tasks at the residue level, we adopt the SMPC pre-training strategy. At the atom level, we employ a random noise strategy. Through a series of structure pre-training tasks, such as position and torsion angle prediction, our model effectively learns residue and atom representations jointly, enabling comprehensive protein modeling at both levels. An overview of our approach is shown in Figure~\ref{fig:FG-prot3d-architecture}.


We design a series of downstream tasks to assess the impact of \name. These tasks include Enzyme Commission (EC) number prediction and Gene Ontology (GO) term prediction, which focus on the global properties of proteins, as well as binding site prediction, which emphasizes protein local properties. Furthermore, we incorporated molecular docking, a task known to rely on atom modeling of proteins, as each atom may interact with small molecules. We incorporate the atom-level representation generated by our model and input it into an existing docking model, subsequently evaluating its effectiveness through performance improvement in molecular docking tasks. Across these tasks, our model, \name, significantly outperforms previous baselines, demonstrating its superior efficacy. In summary, our contributions are as follows:

\begin{enumerate}
\item We present a novel pre-training model \name that learns effective residue and atom representations simultaneously.
\item We propose SMPC strategy to enhance the pre-training task at the residue level, resulting in a significant improvement in performance.
\item \name achieves state-of-the-art results on various downstream tasks, including the molecular docking task. These results demonstrate the efficacy of the protein representations generated by \name.
\end{enumerate}

\section{Related Works}
\textbf{Pre-trained Protein Models}
Pre-training using protein sequences has recently attracted significant attention due to its potential applications. A family of models, which includes ProtTrans~\cite{ProtTrans}, ESM-1b~\cite{rao_msa_2021}, and ESM2~\cite{lin2023evolutionary}, utilize individual protein sequences as input and undergo pre-training by optimizing the masked language model objective. A prompt-based pre-trained protein model~\cite{wang2022multi} has been introduced to guide multi-task pre-training. Additionally, xTrimoPGLM~\cite{chen_xtrimopglm_2023} is proposed to further explore the potential of a unified pre-training strategy.

Recent advancements in structure-based pre-training models have demonstrated significant improvements in performance across a diverse range of downstream tasks.
The prevailing protein pre-training models primarily focus on predicting a diverse range of physical quantities, including torsion angles, angles between edges, alpha carbon positions, and distances, as well as residue types. \cite{zhang2022protein, guo_self-supervised_2022, chen_structure-aware_2023,wang2022multi}.
Contrastive learning has demonstrated exceptional performance in a recent study based on GearNet~\cite{zhang2022protein}.
Additionally, recent attempts for pre-training with protein surfaces have been made~\cite{noauthor_pre-training_2023}. 
Numerous diffusion-based models have been proposed~\cite{huang_data-efficient_2023, zhang2023pre}. For instance, SiamDiff~\cite{zhang2023pre} employs a pre-trained protein encoder through sequence-structure joint diffusion modeling. HotProtein\cite{chen2022hotprotein}, a structure-aware pre-training model, is proposed to improve thermostability prediction. A self-supervised pretraining method is also used to predict compound–protein affinity and contact\cite{you2022cross}.

Although significant progress has been made in the development of residue-level pre-training models, there is a scarcity of focus on atom-level pre-training models for proteins.  Among very few of them, Siamdiff~\cite{zhang2023pre} can function as either a residue or atom-level model.

The GearNet style model~\cite{zhang2022protein, noauthor_pre-training_2023} has demonstrated superior performance compared to other residue-level models~\cite{zhang2022protein, zhang2022protein}. In addition to GearNet, we select Siamdiff~\cite{zhang2023pre} as another primary benchmark, as it has been reported to be the most effective atom-level pre-training model. Together, these two models provide a comprehensive and robust foundation for our analysis and comparison.

\textbf{Residue Level Encoder.} In addition to numerous pre-training models, several studies attempt to encode protein structures in various downstream tasks without pre-training. To leverage structural information, plenty of models using structure or structure+sequence information have been proposed. GVP~\cite{jing_equivariant_2021} iteratively updates the scalar and vector representations of a protein to learn its representation. Additionally, CDConv~\cite{fan2022continuous} employs both irregular and regular approaches to model the geometry and sequence structures. Furthermore, ProNet~\cite{wang2023learning} utilizes torsion angles to capture side-chain positions.

\textbf{Atom Level Encoder.} Because of the importance of side chain atoms, there have also been some all-atom-level encoders proposed recently. IEConv \cite{hermosilla_intrinsic-extrinsic_2021} introduce a novel convolution operator and a hierarchy pooling operator, which facilitated multi-scale protein analysis. FAIR \cite{zhang2023full} attempts to encode atom and residue-level information by employing two separate encoders, without incorporating any interaction between them. 
\section{Method}
\subsection{Vector Aware Bilevel Sparse Attention Network}

\begin{figure*}[ht]
    \centering
    \includegraphics[width=\textwidth]{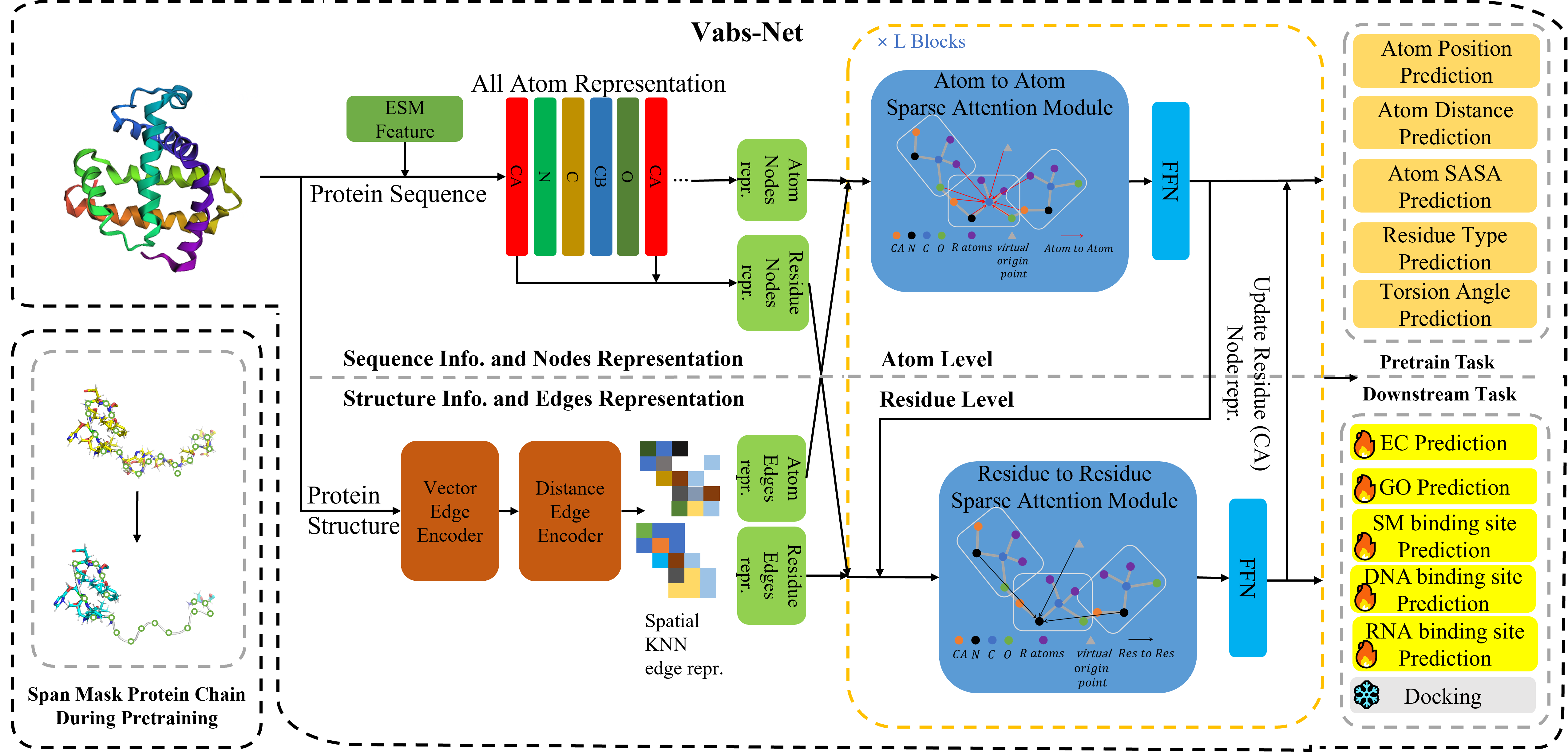}
    \caption{
    An overview of \name architecture. We use atom type, residue type, and preprocessed ESM features to encode atom nodes. Residue nodes share representation with their corresponding alpha carbon. Encoding of edges is through vector edge encoder and distance edge encoder to encode direction and distance of edges. We input node and edge encoding into a two-track sparse attention module. Each track includes a sparse attention module and a feedforward neural network. This module first updates atom representations with the atom-atom track and then updates alpha carbon atom nodes by residue-residue track. In this way, two tracks interact through alpha carbon atom nodes. Finally, representations of nodes and edges are used for various pre-training and downstream tasks. In addition, we show the span mask protein chain strategy on the left. Atom nodes other than alpha carbon are removed in the masked area of the span mask protein chain method during pre-training.
    }
    \label{fig:FG-prot3d-architecture}
\end{figure*}

\textbf{Protein Graph Construction} 
We first construct both residue-level and atom-level graphs. A residue level graph can be defined as  $\mathcal{G}_{res}=(\mathcal{V}_{CA}, \mathcal{E}_{res})$, where $\mathcal{V}_{CA}$ includes all alpha carbon nodes and $\mathcal{E}_{res}$ includes all residue edges. An atom level graph can be defined as $\mathcal{G}_{atom}=(\mathcal{V}, \mathcal{E}_{atom})$, where $\mathcal{V}$ includes all-atom nodes, i.e. $\mathcal{V}_{CA} \subset  
 \mathcal{V}$ and $\mathcal{E}_{atom}$ includes all atom level edges. The construction of edges involves the selection of k nearest neighbors in the space for both residue and atom-level graph nodes. The residue-level graph shares alpha carbon nodes with the atom-level graph. Consequently, the bilevel graph of protein can be written as $\mathcal{G}=(\mathcal{V}, \mathcal{E}_{res}, \mathcal{E}_{atom})$.
The model architecture can be seen in Figure\ref{fig:FG-prot3d-architecture}. This approach facilitates information exchange between atom and residue levels, thereby not only allowing atoms to acquire an expanded receptive field but also enabling residues to obtain detailed structural information.


To comprehensively encapsulate global information on protein, we introduce a virtual origin point positioned at the geometry center. This point is connected with every atom and residue, integrating overall protein representation.

\textbf{Node Encoding} 
In our model, node embeddings incorporate both atom type and residue type. Additionally, we utilize the Large language model (LLM) version ESM~\cite{lin2023evolutionary} to leverage sequence information. The representation of atom nodes can be seen as follows:
$$
\boldsymbol{x} = \textrm{Embeding}(ATOM) + \textrm{Embeding}(RES) + \boldsymbol{W_{E}}\boldsymbol{v}_{E},   
$$
where $ATOM$ and $RES$ stand for the type of atom and residue. $\boldsymbol{W_{E}}$ maps the shape of ESM repr. to node shape.

\textbf{Edge Distance Encoding} 
To encode the distance between atoms or residues, we use the Gaussian kernel.
\begin{align*}
g^k_{i,j} &= \frac{1}{\sigma^k\sqrt{2\pi}} \exp\left(-\frac{1}{2}\left(\frac{\alpha_{i,j}\lVert \boldsymbol{r}_i - \boldsymbol{r}_j \rVert - \mu^k + \beta_{i,j}}{\sigma^k}\right)^2\right),\\
& \qquad\boldsymbol{g}_{i, j} = \mathrm{concat}(g_{ij}^k),\quad k = \{1, 2, \ldots, K\}
\end{align*}
where $\boldsymbol{g}^{k}_{i,j}$ is the k-th Gaussian kernel of the nodes pair $(i, j)$, $K$ is the number of kernels. The input 3D coordinate of the $i$-th atom is represented by $\boldsymbol{r}_i \in \mathbb{R}^3$, and $\alpha_{i,j}$ and $\beta_{i,j}$ are learnable scalars indexed by the pair of node types. $\mu^{k}$ and $\sigma^{k}$ are predefined constants. Specifically, $\mu^{k} = w \times (k - 1) / K$ and $\sigma^{k} = w / K$, where the width $w$ is a hyper-parameter.

\textbf{Edge Vector Encoding}
In previous protein-pretraining modeling, structural information is traditionally encoded using the distance between residues. While this distance-based structural encoding may suffice for residue modeling, encoding distance is not informative enough for atom modeling since the number of atoms is much larger than that of the residue\ref{app:vec}. Encoding edge directions in both the residue local coordinate system and the absolute global coordinate system can alleviate this problem. The residue local coordinate system can be formulated as following steps with three backbone atoms(CA, C, N) in each residue. We use $\boldsymbol{r}$ to indicate the position of atoms and set $\boldsymbol{v}_N=\boldsymbol{r}_N-\boldsymbol{r}_{CA}$ and $\boldsymbol{v}_C=\boldsymbol{r}_C-\boldsymbol{r}_{CA}$ as the vector of edges between N and CA, CA and C in the absolute global coordinate system. More generally,  We define $\boldsymbol{v}_{l}$ and $\boldsymbol{v}_{g}=\boldsymbol{r}_{atom}-\boldsymbol{r}_{CA}$ are the vectors of edges from CA to an atom in the same residue in the local residue coordinate system and global coordinate system, respectively. They serve to encode edge directions in subsequent steps. Then we can get a local coordinate system with a rotational matrix $\boldsymbol{R}$, which is used to transform vectors in global coordinates to local coordinates.
$\boldsymbol{R}$ is constructed with unit vector $\boldsymbol{u},\boldsymbol{v},\boldsymbol{w}$ in XYZ axis of residue local coordinate.
\begin{align*}
\boldsymbol{u} = \frac{\boldsymbol{v}_N - \boldsymbol{v}_C}{\| \boldsymbol{v}_N - \boldsymbol{v}_C \|},\quad \boldsymbol{v} = &\frac{\boldsymbol{v}_N \times \boldsymbol{v}_C}{\| \boldsymbol{v}_N \times \boldsymbol{v}_C \|},\quad \boldsymbol{w} = \boldsymbol{u} \times \boldsymbol{v}, \\
\boldsymbol{R} = [\boldsymbol{u}, \boldsymbol{v}, \boldsymbol{w}]&,\quad \boldsymbol{v}_{l} = \boldsymbol{R}^{\top}\boldsymbol{v}_{g}.
\end{align*}
Empirical evidence~\cite{yifan_input-level_2022} suggests that this higher-dimensional Fourier encoding is more apt for subsequent processing by neural networks. The Fourier encoded representation is defined as:
\begin{align*}
\boldsymbol{\gamma}(\theta)= \bigg[ &\sin\left(2^0 \frac{\pi}{\theta}\right), \cos\left(2^0 \frac{\pi}{\theta}\right), \ldots, \\
&\sin\left(2^{L-1} \frac{\pi}{\theta}\right), \cos\left(2^{L-1} \frac{\pi}{\theta}\right) \bigg].
\end{align*}
Thus we encode the direction of an edge from atom $i$ to atom $j$ by concatenating Fourier encoding angle between edges and xy, yz, and xz planes.
$$
\boldsymbol{f}_{i, j} = \left[\boldsymbol{\gamma}(\phi^{xy}_{ij})\|\boldsymbol{\gamma}(\phi^{yz}_{ij})\|\boldsymbol{\gamma}(\phi^{xz}_{ij})\|\boldsymbol{\gamma}(\varphi^{xy}_{ij})\|\boldsymbol{\gamma}(\varphi^{yz}_{ij})\|\boldsymbol{\gamma}(\varphi^{xz}_{ij}) \right]
$$
We set $\phi^{xy}_{ij}=\arcsin(\boldsymbol{v_g}\cdot \boldsymbol{\hat{z}_g})$ the angle between the vector from node $i$ to node $j$ and the global absolute xy plane, where $\boldsymbol{\hat{z}_g}$ is the unit vector in the z-direction in the global absolute coordinate system. $\varphi^{xy}_{ij}=\arcsin(\boldsymbol{v_l}\cdot \boldsymbol{\hat{z}_l})$ is the angle between the residue local vector from node $i$ to node $j$ and local residue xy plane, where $\boldsymbol{\hat{z}_l}$ is the unit vector in local coordinate. Other angles can be obtained using the same method.

Learnable positional encoding is also applied to encode sequential positions.
As a result, the representation of edge between atoms or residues can be summarized as follows:
$$
\boldsymbol{e}_{i, j} = \boldsymbol{W_g}\boldsymbol{g}_{i, j} + \boldsymbol{W_f}\boldsymbol{f}_{i,j} + \boldsymbol{l}(i-j),
$$
where $\boldsymbol{l}(i-j)$ is the learnable positional encoding based on the sequence distance between two residues. $\boldsymbol{W_g}$ and $\boldsymbol{W_f}$ maps the dimension of distance and direction encoding to dimension of edge.

\textbf{Sparse Attention Module(SAM)} 
Numerous pre-training models for molecules based on Graphormer have achieved state-of-the-art performance in downstream tasks~\cite{zhou_uni-mol_2023}. However, a fully connected Graphormer model can pose significant computational challenges~\cite{lu2023highly}. To tailor Graphormer's attention mechanism to the all-atom level graph, we introduce a Sparse Graph Attention Module aimed at reducing computational load and preventing overfitting in downstream tasks. The sparse attention module is formulated as follows:
$$
a^{l}_{i, j} = \mathrm{softmax}_{j\in\mathcal{N}_i}\left(\dfrac{\boldsymbol{W_Q}\boldsymbol{x}^{l}_i(\boldsymbol{W_K}\boldsymbol{x}^{l}_j)^{\top}}{\sqrt{d_h}} + \boldsymbol{W_B} \boldsymbol{e}_{i, j}\right),
$$
$$
\boldsymbol{x}^{l + 1}_i = \sum_{j\in\mathcal{N}_i}a^{l}_{i, j}\boldsymbol{W_V} \boldsymbol{x}^{l}_j,
$$
where $\boldsymbol{x}^{l}_i$ is the representation of node $i$ in the $l$th layer. $a^{l}_{i, j}$ is the attention weight. Set $\mathcal{N}_i$ includes all nodes connected to node $i$. $\boldsymbol{W_B}$ is a linear layer for bias calculation. $\boldsymbol{e}_{i, j}$ is the edge representation between node $i$ and node $j$. The $h$ dimension in multi-head attention is omitted.

\subsection{Pre-training Tasks}

\textbf{Span Mask Protein Chain(SMPC) Residue Type Prediction} During the pre-training phase, span masking is applied to mask residue types of consecutive biologically meaningful substructures. Specifically, only the alpha carbon atom is retained for the span-masked residues, while all other atoms are eliminated. This aims to prevent information leakage from side chain atoms and other backbone atoms, which could make residue type prediction and torsion angle prediction trivial. We find even backbone atoms can leak residue type to some extent. Subsequently, only alpha carbon representation is employed for residue type prediction. Cross-entropy loss serves as the objective function for predicting residue types.

\textbf{Span Mask Protein Chain(SMPC) Side Chain and Backbone Torsion Angle Prediction} Drawing upon the feature of alpha carbon atoms in SMPC part, \name predicts the cosine and sine components for seven side chain and backbone torsion angle angles. In addition to the L1 norm loss for cosine and sine values, the loss function encompasses a term that ensures the normalization of the sum of squared sine and cosine values to promote accurate angular representation. The loss function for this task is similar to the one used in AlphaFold2 \cite{jumper_highly_2021}.

\textbf{Atom Position and Distance Prediction} 
Apart from the residue level pre-training tasks. To learn atom-level structural information, We randomly choose consecutively residues independent of SMPS. For these residues, we simply add Gaussian noise to atoms in those residues instead of removing atoms. For these atoms, positions are predicted by a movement prediction head.

The position movement prediction head we adopted is similar to the one used in Graphormer-3D~\cite{shi2022benchmarking, lu2023highly}. 
$$
a_{i, j} = \mathrm{softmax}_{j\in\mathcal{N}_i}\left(\dfrac{\boldsymbol{W^{\prime}_Q}\boldsymbol{x}_{i}(\boldsymbol{W^{\prime}_K}\boldsymbol{x}_{j})^{\top}}{\sqrt{d_h}} + \boldsymbol{W^{\prime}_B} \boldsymbol{e}_{i, j}\right),
$$
$$
\boldsymbol{b}_{i, j} = \sum_{j \in \mathcal{N}_i} a_{i, j} (\boldsymbol{r}_i^{N} - \boldsymbol{r}_j^{N}) \boldsymbol{W^{\prime}_V} \boldsymbol{x}_{j},
$$
$$
\boldsymbol{r}_i^P = \boldsymbol{r}_i^{N} + [\boldsymbol{W_{px}}\boldsymbol{b}_{i, j}^x\|\boldsymbol{W_{py}}\boldsymbol{b}_{i, j}^y\|\boldsymbol{W_{pz}}\boldsymbol{b}_{i, j}^z],
$$
$$
\mathcal{L}_{dist} = \dfrac{1}{n^2}\sum_{i \in \varOmega \mid j \in \varOmega}\|d_{ij}^{P} - d_{ij}^{R}\|_1,
$$
$$
\mathcal{L}_{pos} = \dfrac{1}{n}\sum_{i \in \varOmega}\|\boldsymbol{r}_{i}^{P} - \boldsymbol{r}_{i}^{N}\|_2,
$$
where $\boldsymbol{W^{\prime}_Q}, \boldsymbol{W^{\prime}_K}, \boldsymbol{W^{\prime}_V}, \boldsymbol{W^{\prime}_B}$ are projection heads for movement prediction head. $\boldsymbol{r^N_i}$ is the coordinate of node $i$ with noise. $\boldsymbol{r^R_i}$is the real coordinate of node $i$. $\varOmega$ is the set of nodes with noise. $\boldsymbol{x}_i$ and $\boldsymbol{e}_{i,j}$ are representations of node $i$ and representation of edge between node $i$ and node $j$. $d^P_{i, j}=\|\boldsymbol{r}^P_i-\boldsymbol{r}^P_j\|_2$ and $d^R_{i, j}=\|\boldsymbol{r}^R_i-\boldsymbol{r}^R_j\|_2$ are predicted distance and real distance between node $i$ and node $j$. 

\textbf{Solvent Accessible Surface Area(SASA) Prediction} SASA describes the surface area of a biomolecule, such as a protein or nucleic acid, that is accessible to a solvent. It could be used to understand the shape of a protein. We used freesasa~\cite{mitternacht_freesasa_2016} to calculate SASA of each atom. The loss function for SASA prediction is L1 loss.

\section{Experiments}
In this section, we report our experiment setup and results in training and evaluation of our models for pre-training and downstream tasks. Detailed information about downstream task settings can be seen in Appendix \ref{app:base}.
\begin{table*}[htbp]
\vspace{-6pt}
\caption{Model performance on EC numbers and GO terms prediction tasks($F_{max}$). SMPC stands for span mask protein chain.}
\label{tab:ECGO}
\vskip 0.15in
\begin{center}
\begin{small}
\begin{sc}
\begin{tabular}{lp{1.5cm}p{1.5cm}p{1.5cm}p{1.1cm}}
\toprule
Method & BP & MF & CC & EC \\
\midrule
MTL\cite{wang2022multi}     & 0.445 & 0.640 & 0.503 & 0.869 \\
GRADNORM\cite{chen2018gradnorm, wang2022multi}   &  0.466 & 0.643 & 0.504 & 0.874 \\
LM-GVP\cite{wang_lm-gvp_2022} & 0.417 & 0.545 & 0.527 & 0.664 \\
ROTOGRAD\cite{javaloy2022rotograd}    &  0.470 & 0.638 & 0.509 & 0.876 \\
CDConv\cite{fan2022continuous} & 0.453 & 0.654 & 0.479 & 0.820 \\
\hline
PROMPTPRO.\cite{wang2022multi} &  0.495 & 0.677 & 0.551 & 0.888 \\
ESM\_1b\cite{lin2023evolutionary}  & 0.470 & 0.657 &  0.488 & 0.864 \\
GearNet\cite{zhang2022protein}   & 0.490 & 0.650 & 0.486 & 0.874 \\
Siamdiff\cite{zhang2023pre}   & - & - & - & 0.857 \\
GearNet-ESM\cite{zhang2022protein}   & 0.516 & 0.684 & 0.506 & 0.890 \\
SiamDiff-ESM\cite{zhang2023pre}  & - & - & - & 0.897 \\
GearNet-ESM-INR-MC\cite{noauthor_pre-training_2023} & 0.518 & 0.683 &  0.504 & 0.896 \\
\hline
\name-no-SMPC & 0.496 & 0.667 & 0.552 & 0.876 \\
\name & \textbf{0.531} & \textbf{0.695} & \textbf{0.579} & \textbf{0.900} \\

\bottomrule
\end{tabular}
\end{sc}
\end{small}
\end{center}
\vskip -0.1in
\end{table*}

\subsection{Pre-training}
\textbf{Setting} During this pre-training SMPC phase, we mask consecutive spans of the protein residues. The span mask lengths follow a Poisson distribution with a mean ($\lambda$) of 6 and cumulatively makeup 30\% of the protein chain. For those span-masked residues, we only retain alpha carbon atoms. We then construct a protein KNN graph after SMPC.
Also, we randomly sample 30\% of atoms of consecutive residues and add Gaussian noise. The consecutive span of residues also follows the same Poisson distribution with a mean of 6. Gaussian noise is added to atoms with a scale of 0.5\AA. More detailed information about our model can be seen in Appendix \ref{app:detail}.
The pre-training dataset is constructed from the Protein Data Bank and structures predicted by AlphaFold~\cite{jumper_highly_2021}. To obtain high-quality data, structures from the Protein Data Bank Database~\cite{rose2016rcsb} with a resolution greater than 9 are filtered out. Structures from the AlphaFold2 Database with a pLDDT lower than 70 are filtered out. Additionally, MMSeq2~\cite{mirdita_fast_2021} is utilized to cluster the dataset. Finally, we get 163412 clusters from 1.3 million structures to accelerate pre-training by preventing the model from learning repetitive similar structures. During the training process, for each epoch, we randomly sample one structure from each cluster, thus leading to 163412 samples in one epoch. This leads to a more efficient pre-training, where we only need to train 100 epochs (containing only 16w clusters/samples in one epoch) to reach the best performance.


\subsection{Protein Function Prediction}

Enzyme Commission (EC) number prediction involves the anticipation of the EC numbers associated with diverse proteins, delineating their role in catalyzing biochemical reactions. EC numbers are drawn from the third and fourth tiers of the EC tree, resulting in the formation of 538 binary classification tasks~\cite{hermosilla_intrinsic-extrinsic_2021}. Moreover, Gene Ontology (GO) term prediction focuses on determining whether a protein is associated with specific GO terms. These terms categorize proteins into interconnected functional classes within three ontologies: molecular function (MF), biological process (BP), and cellular component (CC).

\textbf{Setting} For EC and GO prediction, we use the same datasets as former researches~\cite{zhang2022protein}. We utilize the protein-level $F_{max}$ to assess performance.

\textbf{Result} The results of our model on BP, MF, CC, and EC can be seen in Table \ref{tab:ECGO}, which clearly shows that our \name model exhibits superior performance in comparison to all baselines with respect to BP, MF, CC, and EC. A substantial enhancement compared to GearNet-ESM demonstrates the efficacy of atom-level encoding. The superior performance over Siamdiff-ESM shows the importance of residue-level encoding and the effectiveness of the proposed SMPC pre-training strategy in learning efficient residue-level representations. 
To further validate SMPC's efficacy, we compared our model with and without SMPC integration during pre-training. The findings demonstrate that incorporating SMPC contributes to effective residue-level representation learning.

\subsection{Protein Binding Site Prediction} 
\begin{table*}[htbp]
\caption{Small Molecule Binding Site Prediction Result in terms of IoU(\%).}
\label{tab:SM}
\vskip 0.15in
\begin{center}
\begin{small}
\begin{sc}
\begin{tabular}{lcp{1cm}p{1.1cm}p{1.1cm}p{1.1cm}p{1.1cm}p{1.3cm}}
\toprule
Method & pre-train  & B277 & DT198 & ASTEX85 & CHEN251 & COACH420 \\
\midrule
FPocket\cite{le_guilloux_fpocket_2009} & $\times$   & 31.5 & 23.2 & 34.1 & 25.4 & 30.0 \\
SiteHound\cite{hernandez2009sitehound} & $\times$  & 36.4 & 23.1 & 38.9 & 29.4 & 34.9 \\
MetaPocket2\cite{macari2019computational} & $\times$  & 37.3 & 25.8 & 37.5 & 32.8 & 37.7 \\
DeepSite\cite{jimenez2017deepsite} & $\times$   & 34.0 & 29.1 & 37.4 & 27.4 & 33.9 \\
P2Rank\cite{krivak2018p2rank} & $\times$     & 49.8 & 38.6 & 47.4 & \textbf{56.5} & 45.3 \\
\hline
ESM2\_150M\cite{lin2023evolutionary}   & $\surd$   & 19.6 & 16.6 & 20.5 & 18.9 & 22.0 \\
GearNet\cite{zhang2022protein}   & $\surd$  & 39.9 & 35.8 & 41.0 & 36.4 & 41.3 \\
Siamdiff\cite{zhang2023pre}   & $\surd$  & 37.7 & 31.0 & 40.7 & 35.3 & 40.3 \\
\name   & $\times$  & 57.7 & 48.6 & 57.8 & 53.2 & 61.4 \\
\name   & $\surd$  & \textbf{60.1} & \textbf{52.0} & \textbf{58.8} & 56.3 & \textbf{64.1} \\
\bottomrule
\end{tabular}
\end{sc}
\end{small}
\end{center}
\vskip -0.1in
\end{table*}
\begin{table}[htbp]
\vskip 0.15in
\caption{DNA Binding Site Prediction Result trained on DNA-573 Train, tested on DNA-129 Test.}
\label{tab:DNA}
\begin{center}
\begin{small}
\begin{sc}
\begin{tabular}{lcp{0.6cm}p{1cm}r}
\toprule
Method & pre-train & auc \\
\midrule
TargetDNA\cite{hu2016predicting}    & $\times$ & 0.825 \\
DNAPred\cite{zhu2019dnapred}   & $\times$ & 0.845 \\
SVMnuc\cite{su2019improving}      & $\times$ & 0.812 \\
COACH-D\cite{wu2018coach}      & $\times$ & 0.761 \\
NucBind\cite{su2019improving}      & $\times$ & 0.797 \\
DNABind\cite{liu2013dnabind}      & $\times$ & 0.858 \\
GraphBind\cite{xia2021graphbind}   & $\times$ & \underline{0.927} \\
\hline
ESM2\_150M\cite{lin2023evolutionary}   & $\surd$ & 0.779 \\
GearNet\cite{zhang2022protein}     & $\surd$ & 0.849 \\
Siamdiff\cite{zhang2023pre}     & $\surd$ & 0.823 \\
\name   & $\times$ & 0.912 \\
\name   & $\surd$ & \textbf{0.940} \\
\bottomrule
\end{tabular}
\end{sc}
\end{small}
\end{center}
\vskip -0.1in
\end{table}

Precise prediction of protein-ligand binding site forms the bedrock of comprehending diverse biological activities and facilitating the design of novel drugs \cite{pei2023fabind}. The binding site prediction task is to predict whether an atom or a residue is a binding site (binary classification task) without inputting ligands. The localization of the binding site relies heavily on the fine-grained local structure.

The prediction of binding sites is hindered by the limited training data, as evident in the constrained sizes of the training sets for DNA and RNA, standing at 573 and 495 samples, respectively~\cite{zhang_us-align_2022}. Traditional methods typically entail an extensive input of features into the model such as atom mass, B-factor, electronic charge, whether it is in a ring, and the van der Waals radius of the atom, among others~\cite{xia2021graphbind}. Hence, our aim is to leverage pre-training methods to learn meaningful atom representations and mitigate the risk of overfitting.

Prior research efforts generally focus on handling single tasks like small molecules or nucleic acid. Comprehensive binding site prediction experiments are made on small molecules, DNA, and RNA to test our model.

\textbf{Setting} We have constructed a large high-quality small molecule binding site dataset, and we use this dataset to train our model and other baselines, which will be open access to the public. For all-atom models, the label of Alpha C is the residue label. In DNA and RNA binding site prediction, we use the AUC for evaluation. Baselines are finetuned using the same setting as their original paper.

In the preparation of our small molecule training dataset, we utilize three distinct datasets: CASF-2016 coreset, PDBBind v2020 refined set, and MOAD. During dataset preparation, we follow the original ligand records in these datasets to extract the protein and ligand components based on the corresponding PDB ID from RCSB. The extracted segments undergo a structural repair process using in-house scripts. For proteins, this process includes the repair of missing residues, replenishment of absent heavy atoms, and addition of hydrogen atoms. In the case of ligands, the process involves repairing bond orders, adding hydrogen atoms, and determining the correct protonation states based on the pocket environment. A comparison of our dataset and frequently used scPDB can be seen in Appendix \ref{app:small_molecule_dataset}. To avoid leakage, MMSeqs2 is used to filter out high protein sequence similarity(similarity above 40\% which is also used in AlphaFold2~\cite{jumper_highly_2021}) with test sets.

Datasets for DNA and RNA are downloaded from the Biolip and Graphbind website. 

To construct the validation set, we used MMSeqs2 to cluster the training set so that the sequence similarity between the validation set and the training set is lower than 40\%. 

\textbf{Result} Tables \ref{tab:SM}, \ref{tab:DNA} and \ref{tab:RNA} provide results of pre-trained models and none pre-trained models on binding site prediction. 
We found that \name outperforms all of the baseline models. 
Baseline models for binding site prediction are typically not pre-trained. Therefore, we compared the best baselines with \name without pre-training, revealing that \name performs competitively with the best baselines when not pre-trained, thereby demonstrating the effectiveness of our backbone. Moreover, incorporating pre-training substantially enhances performance, surpassing other pre-trained models considerably. The comparison between other pre-training models and best baselines without pre-training shows the importance of direction encoding.
\begin{table}[htbp]
\vspace{-6pt}
\caption{RNA Binding Site Prediction Result trained on RNA-495 Train, tested on RNA-117 Test}
\label{tab:RNA}
\vskip 0.15in
\begin{center}
\begin{small}
\begin{sc}
\begin{tabular}{lcccr}
\toprule
Method & pre-train & auc \\
\midrule
RNABind+\cite{walia2014rnabindrplus}    & $\times$ & 0.717 \\
SVMnuc\cite{su2019improving}  & $\times$ & 0.729 \\
COACH-D\cite{wu2018coach}     & $\times$ & 0.663 \\
NucBind\cite{su2019improving}      & $\times$ & 0.715 \\
aaRNA\cite{miao2015large}    & $\times$ & 0.771 \\
NucleicNet\cite{lam2019deep}      & $\times$ & 0.788 \\
GraphBind\cite{xia2021graphbind}  & $\times$ & \underline{0.854} \\
\hline
ESM2\_150M\cite{lin2023evolutionary}   & $\surd$ & 0.699 \\
GearNet\cite{zhang2022protein}    & $\surd$ & 0.778 \\
Siamdiff\cite{zhang2023pre}    & $\surd$ & 0.735 \\
\name   & $\times$ & 0.834 \\
\name   & $\surd$ & \textbf{0.880} \\
\bottomrule
\end{tabular}
\end{sc}
\end{small}
\end{center}
\vspace{-6pt}
\end{table}
\subsection{Molecular Docking}
Predicting the binding structure of a small molecule ligand to a protein, a task known as molecular docking, is crucial for drug design \cite{pei2023fabind}. Equibind~\cite{stark_equibind_2022} and Diffdock~\cite{corso_diffdock_2023} are the two most commonly used models for this purpose. However, the diffusion-based Diffdock requires substantial computational resources; thus, we employ Equibind \cite{stark_equibind_2022} to evaluate the efficacy of features from our model. Equibind is trained using features extracted from various structural pre-training models. 
To be more specific, we trained 4 distinct Equibind models. The first one is Equibind itself. For the other three models, we add features from GearNet~(pre-trained by multiview contrast learning), Siamdiff~(atom-level), and our \name to the node representation of Equibind. Those features are pre-processed before training.
It is also important to note that Equibind is a residue-level model. To optimize the utilization of atom-level features from both the Siamdiff model and our own, we implement a single attention layer to aggregate atom features to the residue node representation of Equibind.

\textbf{Setting} For molecular docking, due to time constraints, we adopt a faster and more stable version of Equibind, which can also be seen in the official repository. More detail is in Appendix \ref{app:Equibind}. To conduct a fair comparison between our model and other pre-training models, we incorporate ESM features into all the models under consideration. Both Equibind and its variant enhanced with pre-trained features are trained for 300 epochs. We use PDBBind as a dataset, which is also used by Equibind~\cite{stark_equibind_2022}.

\textbf{Result} The effects of incorporating features from various structural protein pre-training models on the molecular docking task can be observed in Table \ref{tab:docking}. These findings demonstrate that the atom and residue representation acquired by our model outperforms those of other pre-training models. Incorporating atom-level encoding, our model demonstrates superior performance compared to GearNet. The improved results of both our model and Siamdiff over GearNet further emphasize the significance of atom-level encoding. Owing to the SMPC pretraining method, which aids in learning a more refined residue-level representation, our model surpasses Siamdiff in performance.
\begin{table}[htbp]
\vspace{-8pt}
\caption{Molecular docking results with Equibind.}
\label{tab:docking}
\begin{center}
\begin{small}
\begin{sc}
\begin{tabular}{lcccr}
\toprule
Method & \makecell[c]{LIGAND \\ RMSD}  & \makecell[c]{CENTROID \\ RMSD} \\
\midrule
Equibind    & 8.91 & 6.02 \\
Equibind+GearNet    & 7.82 & 5.34 \\
Equibind+Siamdiff   & 7.75 & 5.11 \\
Equibind+\name     & \textbf{7.23} & \textbf{4.11} \\
\bottomrule
\end{tabular}
\end{sc}
\end{small}
\end{center}
\vspace{-8pt}
\end{table}
\subsection{Ablation Studies}
To analyze the effect of different components, we choose a protein function prediction task(EC) and a binding site prediction task(small molecule(SM)) for the ablation study.

We investigate different pre-training strategies and model configurations, with the findings presented in Table \ref{tab:ablation}. (1) A comparison between models No.0 and No.1 demonstrates the significance of atom-level encoding, particularly in binding site prediction tasks where side chain atoms play a crucial role. (2) Upon comparing models No.0 and No.2, it becomes clear that residue-level encoding is also important for enhancing the receptive field and capturing residue-level representations. (3) Furthermore, the comparison between models No.0 and No.3 confirms the effectiveness of our proposed SMPC pre-training method in enhancing residue-level representations. (4) The notable decrease in performance observed after removing the vector edge encoder (No.3 and No.4) underscores the importance of direction encoding, rather than solely encoding distance. (5) While leveraging ESM can enhance performance (No.5 and No.3), the impact is not as significant as that of the vector edge encoder. (6) Increasing the KNN parameter does not yield a substantial improvement in the model's performance (No.6 and No.3), possibly due to KNN=30 being adequate for capturing most structural motifs~\cite{tateno1997evolutionary}. (7) Our pre-training strategy significantly enhances the model's performance, as evidenced by the comparisons between No.7 and No.0. (8) To demonstrate the effectiveness of our Sparse Attention Module, we replaced it with the AttMLP module used in PiFold \cite{gao2022pifold} in experiment No. 8. The comparison between No.0 and No.8 clearly illustrates the superior performance of our Sparse Attention Module. (9) The results of ablation studies No.9 and No.10 indicate that neither randomly masking protein chains (not necessarily consecutive) nor only masking side chain atoms (while retaining backbone atoms) can completely prevent information leakage.

\begin{table}[htbp]
\vspace{-10pt}
\caption{Ablation studies of our model in small molecule binding site prediction task(COACH420 test set) and EC prediction task.}
\label{tab:ablation}
\vskip 0.15in
\begin{center}
\begin{small}
\begin{sc}
\begin{tabular}{p{0.3cm}lcccr}
\toprule
No. & Method & SM & EC \\
\midrule
0 & \name    & \textbf{64.1} & \textbf{0.900} \\
\hline
1 & No Atom level    & 56.3 & 0.896 \\
2 & No Residue level    & 62.2 & 0.886 \\
3 & No SMPC    & 63.0 & 0.876 \\
4 & No SMPC,Vector Edge Encoder   & 60.9 & 0.874 \\
5 & No SMPC,ESM     & 61.9 & 0.867 \\
6 & No SMPC,knn=90      & 62.7 & 0.876 \\
7 & No pre-train      & 61.4 & 0.826 \\
8 & SAM$\rightarrow$AttMLP\cite{gao2022pifold}      & 62.3 & 0.881 \\
9 & Span mask side chain atoms    & 63.6 & 0.892 \\
10 & Randomly mask    & 63.3 & 0.884 \\
\bottomrule
\end{tabular}
\end{sc}
\end{small}
\end{center}
\vspace{-10pt}
\end{table}
\section{Limitation}
Compared to the sequential pre-training dataset (1103M in xTrimoPGLM), our structural pre-training dataset (130K) is much smaller. The AlphaFold database now contains over 100 million structures. Also, our pre-training is limited to single-chain pre-training, neglecting interaction between chains. In our subsequent efforts, we seek to tackle the limitations above.
\section{Conclusion}
In this study, we introduce the \name model with the span mask strategy 3D protein chains pre-training technique, aiming to learn atom-level representation and improve residue-level representation. We have conducted extensive experiments on various task types to assess the effectiveness of our \name model and the span mask protein chains pre-training approach. Our \name model demonstrates superior performance, surpassing previous state-of-the-art models. In the subsequent phase of our research, we plan to enhance the integration of sequence and structural features.
\newpage

\section*{Acknowledgments}
This study was supported by the National Key Research and Development Program of China (2022YFA1004304).

\section*{Impact Statement}
This paper presents work whose goal is to advance the field of Machine Learning. There are many potential societal consequences of our work, none of which we feel must be
specifically highlighted here.

\bibliography{example_paper}
\bibliographystyle{icml2024}

\newpage
\appendix
\onecolumn
\section{Small Molecule Dataset.}
\label{app:small_molecule_dataset}
During the construction of our dataset, after eliminating systems with unsuccessful protein preparation, ligand bond order repair failures, and high similarity records, our dataset comprises a total of 22995 samples for training and 1006 samples for validation.

scPDB seems to be the most popular small molecule binding site dataset currently. Table \ref{tab:scPDB} shows the performance of pre-training baselines and our model on scPDB and our dataset. Our dataset containing 22995 samples is also significantly larger than scPDB which only contains 5564 samples.
\begin{table*}[htbp]
\caption{Small Molecule Binding Site Prediction Result in terms of IoU(\%).}
\label{tab:scPDB}
\vskip 0.15in
\begin{center}
\begin{small}
\begin{sc}
\begin{tabular}{lcp{1cm}p{1cm}p{1cm}p{1cm}p{1cm}p{1cm}}
\toprule
Dataset & Method & ave. & B277 & DT198 & ASTEX85 & CHEN251 & COACH420 \\
\midrule
\multirow{4}{*}{scPDB}& ESM2\_150M  & 18.4 & 18.8 & 15.6 & 17.2 & 18.5 & 21.8 \\
& GearNet   & 29.6 & 28.8 & 26.2 & 31.2 & 29.0 & 32.8 \\
& Siamdiff\_atom & 26.6 & 25.2 & 21.6 & 28.1 & 27.8 & 30.1 \\
& \name    & \textbf{54.8} & \textbf{57.1} & \textbf{48.8} & \textbf{55.1} & \textbf{52.8} & \textbf{60.0} \\
\hline
\multirow{4}{*}{Our Dataset}  & ESM2\_150M  & 19.5 & 19.6 & 16.6 & 20.5 & 18.9 & 22.0 \\
& GearNe  & 38.9 & 39.9 & 35.8 & 41.0 & 36.4 & 41.3 \\
& Siamdiff & 37.0 & 37.7 & 31.0 & 40.7 & 35.3 & 40.3 \\
& \name  & \textbf{58.3}  & \textbf{60.1} & \textbf{52.0} & \textbf{58.8} & \textbf{56.3} & \textbf{64.1} \\
\bottomrule
\end{tabular}
\end{sc}
\end{small}
\end{center}
\vskip -0.1in
\end{table*}

The IoU results of prediction for the binding site of small molecules can be seen in the following table. We will add these results to our appendix. Results show that IoU drops slightly.
\begin{table*}[htbp]
\caption{Results of small molecules vary with the number of residues.}
\label{tab:scPDB}
\vskip 0.15in
\begin{center}
\begin{small}
\begin{sc}
\begin{tabular}{lcp{1cm}p{1cm}p{1cm}p{1cm}p{1cm}p{1cm}}
\toprule
number of residues & 0-200	&200-399	&400-599&	600-799&	800-999 \\
\midrule
Average IoU& 0.663&	0.675&	0.612&	0.501&	0.614 \\
\bottomrule
\end{tabular}
\end{sc}
\end{small}
\end{center}
\vskip -0.1in
\end{table*}

\section{Vector Encoding.}
\label{app:vec}
In most protein pretraining models, distance is frequently utilized to encode protein structure. Theoretically, given KNN=30 (as detailed in our paper) as input, we should be capable of reconstructing a protein chain's configuration (referring to the protein's shape or the 3D coordinates of its atoms within a specific coordinate system.). Multidimensional scaling represents a widely employed algorithm for translating distances between each pair of n objects in a set into a configuration of n points, which are then mapped into an abstract Cartesian space. The computational complexity of this process is $O(N^3)$, where N represents the number of points. Consequently, as the number of atom nodes increases, reconstructing the configuration of a protein becomes more challenging. Notably, the number of atoms is typically eight times greater (based on our pretraining dataset) than the number of residues, which suggests that distance may not be sufficient to encode all-atom configuration. This motivates us to add edge direction features to edges.

\section{More Detailed Information About Model Architecture.}

Our model, not being SO(3) equivariant, requires additional steps to ensure robustness to rotation. To achieve this, we randomly rotate each sampled chain from the dataset and translate its center to the coordinate origin. 

We try to achieve SO(3) equivariance in our model through the following modifications: (1) eliminating the embedding of the edge vector in the global coordinate system, and (2) removing the virtual origin point, as it lacks a relative coordinate system. To pre-train a SO(3) equivalent pretraining model, we also modify the movement head module to make it SO(3) equivalent. We find that making our model SO(3) equivalent can impair the performance of our model as shown in Table \ref{tab:SO3}.

\begin{table}[htbp]
\vspace{-10pt}
\caption{Performance of \name with or without SO(3) equivariance.}
\label{tab:SO3}
\vskip 0.15in
\begin{center}
\begin{small}
\begin{sc}
\begin{tabular}{lcccr}
\toprule
Method & SM & EC \\
\midrule
\name    & \textbf{64.1} & \textbf{0.900} \\
\hline
No SMPC    & 63.0 & 0.876 \\
No SMPC SO(3)      & 60.9 & 0.874 \\
\bottomrule
\end{tabular}
\end{sc}
\end{small}
\end{center}
\vspace{-10pt}
\end{table}

In addition, the ESM model of size 650M is used for pre-training and downstream finetuning.

Throughout the fine-tuning process, with the exception of molecular docking, all parameters remained unfrozen. Additionally, in conventional practice, two evolutionary conversation profiles (PSSM and HMM) are traditionally utilized in DNA and RNA binding site prediction~\cite{xia2021graphbind}. A more comprehensive description of the model is provided in Table \ref{tab:param}. The sizes of DNA and RNA molecules are considerably larger than those of small molecules, necessitating the selection of a k-nearest neighbor (knn) value of 90 to encompass long-range information. In contrast, the pre-training model maintains a knn value of 30. Due to the restricted size of the training set for DNA and RNA binding site prediction, we utilized a model with 256-node dimensions.
\label{app:detail}

\begin{table}[H]
\caption{Model parameters.}
\label{tab:param}
\vskip 0.15in
\begin{center}
\begin{small}
\begin{sc}
\begin{tabular}{lcccr}
\toprule
Method  & DNA,RNA binding site & EC,GO & small molecule binding site\\
\midrule
layers    & 12 & 12 & 12  \\
Node dim.    & 256 & 768 & 768 \\
Edge dim.   & 128 & 128 & 128  \\
ffn dim.    & 512 & 768 & 768  \\
batch size      & 32 & 32 &  64 \\
knn      & 90 & 30 & 30 \\
total step     & 50k & 200k & 100k \\
warmup step     & 5k & 5k & 5k \\
learning rate     & 1e-5 & 5e-5 & 1e-5 \\
Optimizer     & Adam & Adam & Adam \\
\bottomrule
\end{tabular}
\end{sc}
\end{small}
\end{center}
\vskip -0.1in
\end{table}
\section{More Detailed Information About Downstream Task Experiments.} 
\label{app:base}
In all downstream tasks, we introduce random Gaussian noise at a scale of 0.5{\AA} to 20\% of the residues in order to prevent overfitting and enhance the robustness of our model.

We compare our model with two important pre-trained models. The first one is the well-known residue-level GearNet pre-trained by multiview contrast learning. The second one is the all-atom-level Siamdiff. We used the checkpoints downloaded from their respective GitHub pages and employed the original configurations to obtain results for small molecules, DNA, and RNA binding sites tasks.

Notably, all of these models are trained using 16 Tesla A100 GPUs.
\section{More Detailed Information About Equibind Training.} 
\label{app:Equibind}

A faster and more stable version of Equibind in the official repository features 5 layers, 20 attention heads, and 1e-5 weight decay as opposed to the original version using 8 layers, 30 attention heads, and 1e-4 weight decay.

\section{Computational Efficiency}
All training experiments are conducted with 16 Tesla A100. For the downstream task, we choose small molecule binding site prediction. All inference experiments are conducted with 8 V100 on COACH420 test set(409 samples).

\begin{table}[H]
\caption{Time Spent to Pre-train, Finetune, Inference Our All-atom Version and Residue-level Version vabs-net.}
\label{tab:param}
\vskip 0.15in
\begin{center}
\begin{small}
\begin{sc}
\begin{tabular}{lcccccr}
\toprule
Settings  & Atom PT & Residue PT & Atom FT & Residue FT	& Atom Infer & Residue Infer\\
\midrule
Time    & 4d16h &	2d2h	&8h&	3.5h&	22s&	7s  \\
Epochs  & 190& 190  & 50 & 50& 1 & 1 \\
\bottomrule
\end{tabular}
\end{sc}
\end{small}
\end{center}
\vskip -0.1in
\end{table}


\end{document}